\documentclass[10pt,twocolumn,letterpaper]{article}

\usepackage{iccv}
\usepackage{times}
\usepackage{epsfig}
\usepackage{graphicx}
\usepackage{amsmath}
\usepackage{amssymb}
\usepackage{booktabs}
\usepackage{array}
\usepackage{placeins}
\usepackage{hhline}
\usepackage{multirow,makecell}

\usepackage[pagebackref=true,breaklinks=true,colorlinks,bookmarks=false]{hyperref}

\iccvfinalcopy 

\def\iccvPaperID{2667} 

\newcommand{\zhanglin}[1]{#1}

\ificcvfinal\fi

\begin{document}

\title{LMR: A Large-Scale Multi-Reference Dataset for Reference-based Super-Resolution}

\author{Lin Zhang\thanks{Work done during an internship at Baidu Inc.}\\
CASIA
\and
Xin Li \\
Baidu Inc.
\and
Dongliang He \\
Baidu Inc.
\and
Fu Li \\
Baidu Inc.
\and
Errui Ding \\
Baidu Inc.
\and
Zhaoxiang Zhang \\
CASIA
}


\maketitle
\ificcvfinal\thispagestyle{empty}\fi

\begin{abstract}

It is widely agreed that reference-based super-resolution (RefSR) achieves superior results by referring to similar high quality images, compared to single image super-resolution (SISR). Intuitively, the more references, the better performance. 
However, previous RefSR methods have all focused on single-reference image training, while multiple reference images are often available in testing or practical applications.
The root cause of such training-testing mismatch is the absence of publicly available multi-reference SR training datasets, which greatly hinders research efforts on multi-reference super-resolution.
To this end, we construct a large-scale, multi-reference super-resolution dataset, named \textbf{LMR}.
It contains 112,142 groups of 300$\times$300 training images, which is 10$\times$ of the existing largest RefSR dataset. The image size is also much larger. More importantly, each group is equipped with 5 reference images with different similarity levels. 
Furthermore, we propose a new baseline method for multi-reference super-resolution: \textbf{MRefSR},
including a \textbf{M}ulti-Reference \textbf{A}ttention \textbf{M}odule (MAM) for feature fusion of an arbitrary number of reference images, and a \textbf{S}patial \textbf{A}ware \textbf{F}iltering \textbf{M}odule (SAFM) for the fused feature selection.
The proposed MRefSR achieves significant improvements over state-of-the-art approaches on both quantitative and qualitative evaluations. \url{https://github.com/wdmwhh/MRefSR}
\end{abstract}
\section{Introduction}

\begin{figure}[t]
	\centering
	\includegraphics[height=5.5cm]{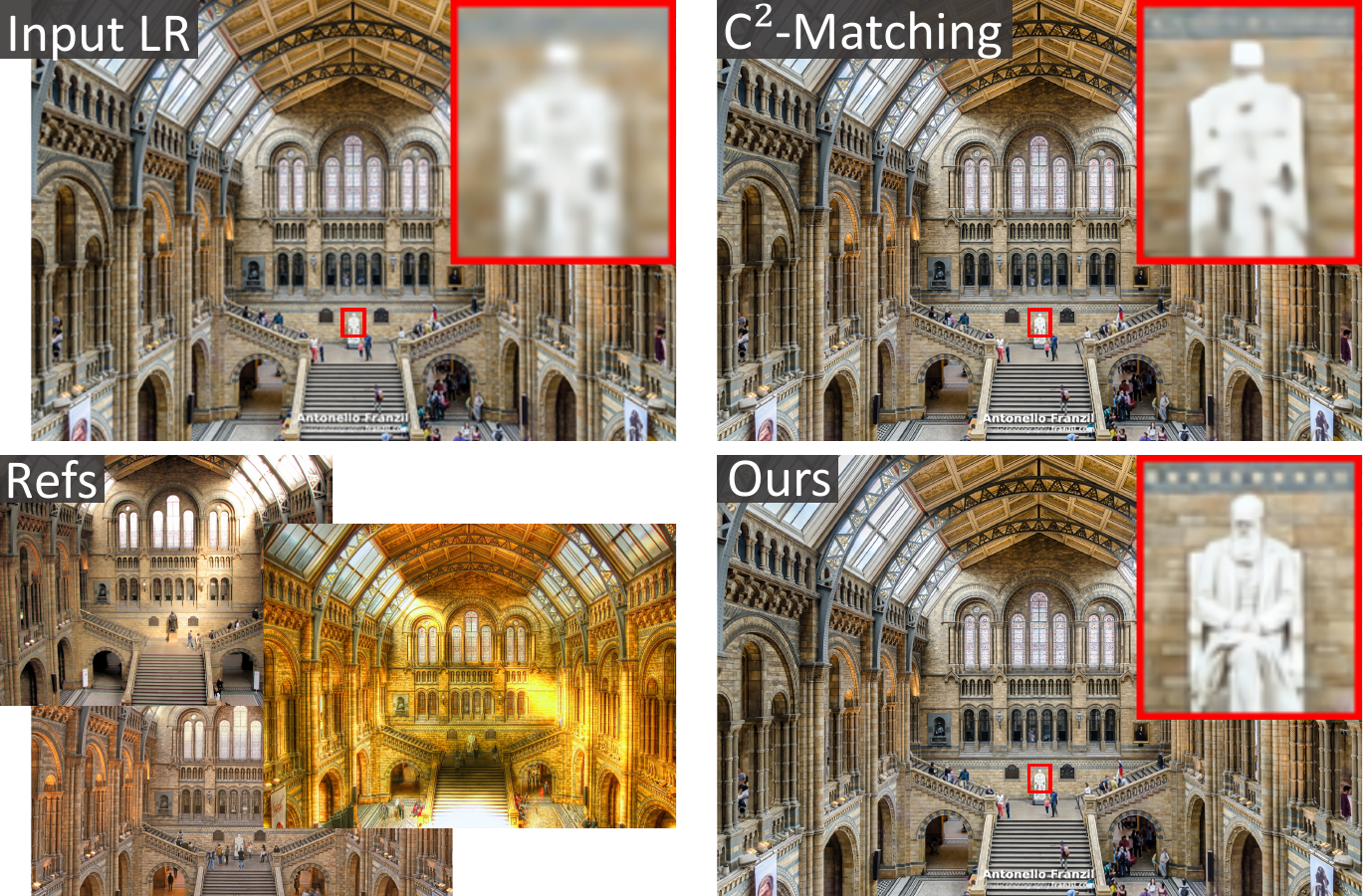}
	\caption{Visual comparison of single-reference training RefSR method $C^2$-Matching \cite{jiang2021robust} and our multi-reference training MRefSR. Our MRefSR can more fully utilize arbitrary number of multiple reference images to achieve the best results. This figure is best viewed by zoom-in.}
	\label{fig:mref_demo}
\end{figure}

Single image super-resolution (SISR) is to restore a degraded low-resolution (LR) image to a texture-realistic high-resolution (HR) image~\cite{irani1991improving}. 
SISR has a wide range of applications in surveillance~\cite{zhang2010super}, astronomy~\cite{holden2011daostorm}, medical imaging~\cite{greenspan2009super}, film and television~\cite{patti1997superresolution,kim20182x}, and other industries~\cite{sevom2018360,zhang2018super,wei2019fpga}. 
With the development of deep learning, SISR has made great progress over these years~\cite{dong2014learning,dong2015image,ledig2017photo,lim2017enhanced,johnson2016perceptual,zhang2018image,wang2018esrgan,liu2018non,dai2019second,mei2021image}. 
Compared with SISR, reference-based super-resolution (RefSR) can leverage textures from additional similar HR reference images, so it often achieves better performance.

Because of promising results shown by recent RefSR methods~\cite{sun2012super,yue2013landmark,zheng2018crossnet,zhang2019image,shim2020robust,tan2020crossnet++, jiang2021robust, lu2021masa, xia2022coarse}, it attracts more and more research interest. 
However, all these previous RefSR methods have focused on using a single reference image for training, but there are often multiple reference images available for testing or practical applications. 
To the best of our knowledge, the only RefSR training dataset currently available is CUFED5~\cite{zhang2019image,wang2016event}, which has only 11,871 image pairs with a small resolution of 160$\times$160. More importantly, there is only one reference image for each LR input image. 
However, in practical applications, multiple reference images are often encountered.
For example, testing set of CUFED5 has 126 input images and each has 5 reference images with different similarity levels.
Similarly, we can also easily find multiple reference images for any real test case. 
Due to the limitation of the only available training dataset, previous RefSR methods do not make good use of multiple reference images in testing or practical applications.
The previous RefSR methods usually stitch together several reference images to get a large resolution image as one reference image to fit the models trained with only one reference image. 
Nevertheless, if the resolution of the reference images is too large, this way of testing will exhaust the GPU memory. \zhanglin{Furthermore, the relationship among multiple reference images is not modeled effectively.} So this is certainly much worse than a method designed specifically for multiple reference images. 
Therefore, a multi-reference RefSR training dataset and a simple but effective multi-reference RefSR method are needed.

In this paper, we propose a large-scale, multi-reference RefSR dataset, named LMR. 
The training set of LMR consists of 112,142 groups of 300$\times$300 training images, each group containing 5 reference images of different similarity levels. 
LMR training dataset has 10 times images compared to CUFED5 and the image size is also much larger. Such a sufficiently large training dataset will be beneficial for improving the generalization ability of models.
We believe this training dataset will greatly facilitate the RefSR research as it is the first RefSR training dataset with multiple reference images.
Meanwhile, the testing set of LMR has 142 groups of images and each group with 2$\sim$6 reference images. The side length of the testing images ranges from 800 to 1600.

With the help of LMR, we propose a new RefSR baseline method for multiple reference RefSR, named MRefSR. 
First, we develop a \textbf{M}ulti-Reference \textbf{A}ttention \textbf{M}odule (MAM) for feature fusion from an arbitrary number of reference images. 
We treat the LR input feature as query, and candidate keys and values are generated from the aligned reference features corresponding to different reference images. 
Then, attention across different aligned reference features is conducted to fuse features from different reference images.
Second, since not all LR feature points can well match the reference features, we use \textbf{S}patial \textbf{A}ware \textbf{F}iltering \textbf{M}odule (SAFM) for fused feature selection. As shown in Figure~\ref{fig:mref_demo}, our MRefSR effectively utilizes information from multiple reference images to produce visually pleasing details.
In summary, our contributions are threefold: 
\begin{itemize}
\item We contribute the first multi-reference RefSR dataset, named LMR, which contains 112,142 groups of 300$\times$300 training images and each group has 5 reference images for the input image. This dataset will enable RefSR research from single-reference to multi-reference images and largely promote the development of the RefSR research field.
\item We propose a novel multi-reference baseline RefSR method MRefSR, using a multi-reference attention module for feature fusion of an arbitrary number of reference images, and a spatial aware filtering module for the fused feature selection. \zhanglin{Our method effectively learns the relationship among multiple references and makes the best use of them, this is also thanks to the multi-reference dataset LRM.} 
\item We conduct extensive experiments which demonstrate the superiority of the proposed LMR and the potential of multi-reference RefSR methods. Our method achieves significant improvements over state-of-the-art approaches on both quantitative and qualitative evaluations.
\end{itemize}

\section{Related Work}
\subsection{Reference-based Image Super-Resolution}
RefSR is gradually becoming an emerging research field. Compared with SISR, RefSR is more advantageous because it can utilize the information of additional HR reference images with similar contents. 
SRNTT~\cite{zhang2019image} proposed an end-to-end network structure that performs multi-scale adaptive texture transfer from the reference image to recover the SR image. 
Subsequently, TTSR~\cite{yang2020learning} applied a cross-scale feature integration method to merge multi-scale reference features. 
MASA~\cite{lu2021masa} designed a coarse-to-fine patch matching scheme to reduce the computational complexity. 
Consequently, $C^2$-Matching~\cite{jiang2021robust} got more accurate pre-offsets of reference features to LR features by a teacher-student correlation distillation and a dynamic DCN~\cite{dai2017deformable,zhu2019deformable} aggregation module.  AMSA~\cite{xia2022coarse} made an incremental extension of $C^2$-Matching by introducing multi-scale aggregation and coarse-to-fine patch matching.  Huang \etal~\cite{huang2022task} also used the $C^2$-Matching model, but added an additional SISR network to decouple the texture transfer and the super-resolution, 
which made the network parameters much larger and the inference much slower. \zhanglin{Recently, RRSR~\cite{zhang2022rrsr} and DATSR~\cite{cao2022reference} also introduce reciprocal learning and transformers to boost the performance.}
Although previous methods have made great progress, all of the above methods focus on research exploration using only one single reference image due to the limitation of the only available training dataset, CUFED5.
\subsection{RefSR Datasets}
\begin{figure*}[!t]
    \centering
	\begin{tabular}{p{0.152\textwidth}<{\centering}p{0.128\textwidth}<{\centering}p{0.145\textwidth}<{\centering}p{0.140\textwidth}<{\centering}p{0.150\textwidth}<{\centering}p{0.135\textwidth}<{\centering}}
		Target & Ref-1 & Ref-2 & Ref-3 & Ref-4 & Ref-5 \\
	\end{tabular}
	\begin{center}
	    \includegraphics[width=0.985\textwidth]{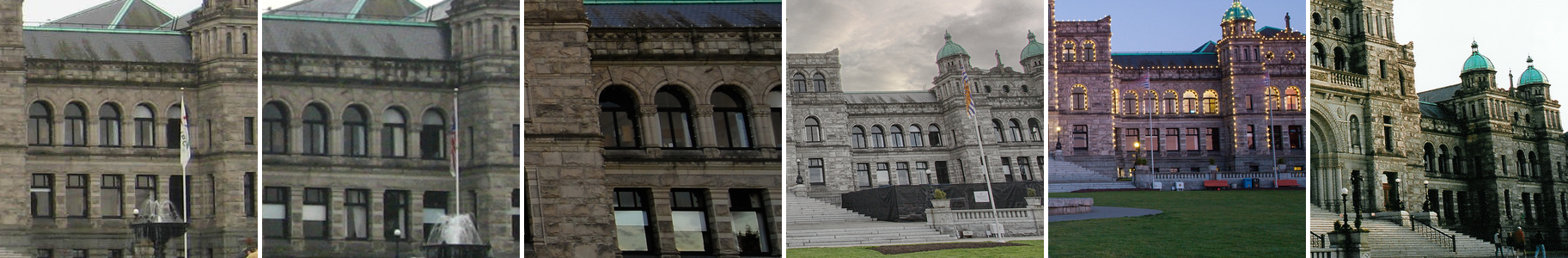}
	    \vskip 3pt
	    \includegraphics[width=0.985\textwidth]{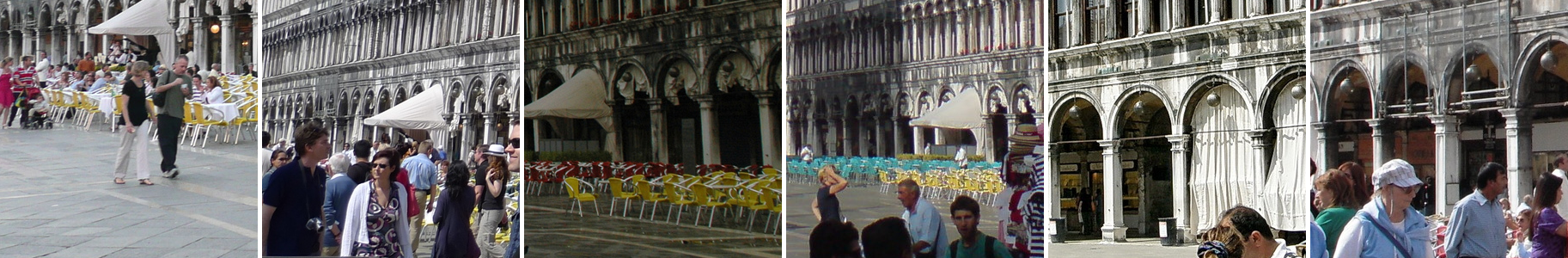}
	\end{center}
    \caption{Two groups of sample images from our LMR training dataset. From left to right, there is one target image, one high-similarity (H) reference image, two medium-similarity (M) reference images, and two low-similarity (L) reference images.}

    \label{fig:samples}
\end{figure*}
To the best of our knowledge, there are five datasets commonly used in RefSR research: Sun80~\cite{sun2012super}, Urban100~\cite{huang2015single}, Manga109~\cite{matsui2017sketch}, WR-SR~\cite{jiang2021robust} and CUFED5~\cite{zhang2019image,wang2016event}. 
However, the first four are all testing sets. 
The Sun80 dataset contains 80 natural images, each with 20 web-search reference images, but these reference images are not very similar to the corresponding LR input, so it is not suitable as a testing set for RefSR.
The Urban100 dataset contains 100 building images, lacking references. Because of self-similarity in the building image, the corresponding LR image is usually treated as the reference image. 
The Manga109 dataset contains 109 manga images without references. Since all the images in Manga109 are the same category (manga cover), the previous methods randomly use one HR image in the dataset as a reference image. 
The WR-SR dataset with more diverse categories, contains 80 image pairs, each target image accompanied by a web-searching reference image. 
CUFED5~\cite{zhang2019image,wang2016event} is the only dataset with a training set, which has 11,871 image pairs with a small resolution of 160$\times$160 and only one reference image for the LR input in each image pair.
CUFED5 testing set has 126 input images and each has 5 reference images with different similarity levels. Recently, Wang \etal~\cite{wang2021DCSR} proposed a new dataset named CameraFusion for dual-camera super-resolution with 131 training image pairs and 15 testing image pairs. However, the image pairs captured by the dual-camera is too ideal for the RefSR task, and the number of dataset is too small. In this paper, to better meet the demands of RefSR research, we propose LMR, a large-scale multi-reference RefSR dataset.

\section{Approach}

In this section, we first introduce the proposed \textbf{L}arge-scale \textbf{M}ulti-reference \textbf{R}efSR dataset \textbf{LMR} in Sec.~\ref{sec:construct_RMD}. Subsequently, we detail a new baseline RefSR method \textbf{MRefSR} using multiple references in Sec.~\ref{sec:mrsa}.

\subsection{Construction of LMR}
\label{sec:construct_RMD}

\begin{figure*}[t]
	\centering
	\includegraphics[width=0.985\textwidth]{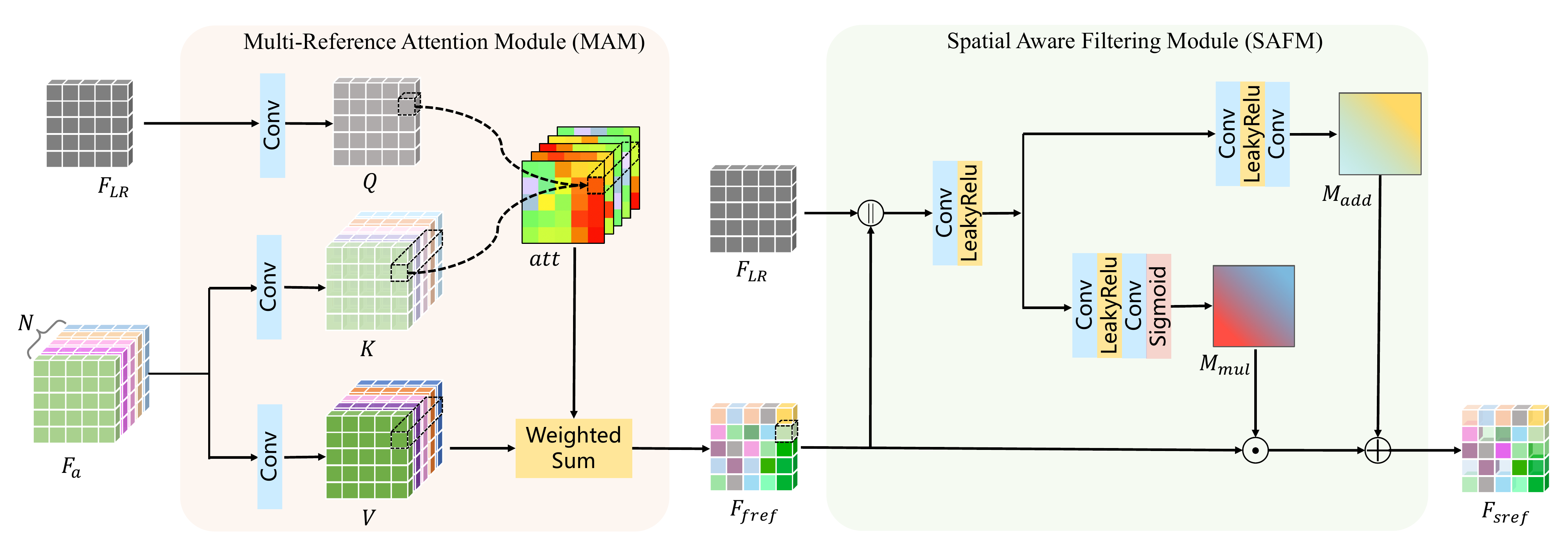}
	\caption{The proposed Multi-Reference Attention Module (left) for the multi-reference feature fusion and the Spatial Aware Filtering Module (right) for the fused feature selection. Both modules perform pixel-wise functions.}
	\label{fig:Method}
\end{figure*}

The MegaDepth~\cite{li2018megadepth} dataset was originally proposed for single-view depth prediction. They used a large number of Internet images from overlapping viewpoints to obtain the dense depth by COLMAP, a state-of-the-art SfM system~\cite{schonberger2016structure} (for reconstructing camera poses and sparse point clouds) and MVS system~\cite{schonberger2016pixelwise} (for generating dense depth maps). The generated dense depth maps of the COLMAP are used as the supervised targets for single-view depth prediction model training. 
MegaDepth contains 1,070,468 internet photos of landmarks around the world and reconstructs 196 3D landmark models from these photos. Each photo of the same landmark varies widely in viewpoint, scene extent, and focused buildings.
The scene of finding Internet images from overlapping viewpoints for 3D reconstruction is very similar to finding reference images for target images to do reference-based super-resolution. 
Inspired by this, the image groups in the off-the-shelf MegaDepth dataset are very suitable for making a RefSR dataset. Consequently, we propose a new large-scale multi-reference RefSR dataset, dubbed LMR.

To construct the LMR training image patch groups, we first perform the following preprocessing steps on the original MegaDepth dataset to obtain similar image pairs.

\begin{itemize}
\item Firstly, the PSNR of the target image and the candidate reference images should be lower than $30$dB to filter duplicate images.
\item Secondly, the candidate reference images and the target image should have some similar contents, and we achieve this filtering by controlling the overlap ratio $R_{olp}$ of matched keypoints in the sparse 3D point clouds.
\item Thirdly, the size ratio $R_{s}$ of the same object in the reference image and the target image cannot be too small, otherwise the reference image cannot provide enough detailed texture information.
\end{itemize}
We calculate $R_{s}$ and $R_{olp}$ following the existing code in D2-Net~\cite{dusmanu2019d2}, a method for image matching and 3D reconstruction.

Further, we define three similarity levels for these image pairs, that is high similarity ($H$), medium similarity ($M$) and low similarity ($L$). 
A image pair is categorized as $H$ if the overlap ratio $R_{olp}$ is greater than $30\%$ and the size ratio $R_{s}$ is larger than $0.9$, $M$ if $R_{olp}$ is greater than $10\%$ and the $R_{s}$ is larger than $0.66$, otherwise $L$.

Through the above operations, we can obtain a large number of image groups, each containing one target image and multiple reference images. However, due to GPU memory limitation, it is often not possible to use the entire large image to train the network. For SISR, it is common to randomly crop a patch from the image for training. 
While in the case of RefSR, it is better to crop corresponding patches with similar contents in the reference images and the target image, e.g. CUFED5 cropped 11,871 paired 160$\times$160 patches as the training set. 
For the multi-reference dataset LMR, we first randomly crop a patch from the target image. 
\zhanglin{Then, we map the center point of the cropped patch into 3D sparse point cloud and pick up 5 keypoints near the mapped point, which are from 5 reference images with different similarities (one H, two M, two L). Next, we take the selected keypoints as centers and crop the corresponding patches.} 
In this way, we collect a total of 112,142 groups of 300$\times$300 patches as the training set, which is ten times larger than CUFED5 and the image size is much larger too. More importantly, each group has 5 reference image patches of different similarities. Some representative samples are presented in Figure~\ref{fig:samples}. As shown in Sec.~ \ref{experiments}, the model trained on the LMR dataset shows good generalization performance on other RefSR datasets, demonstrating the effectiveness of the LMR.

In addition to the LMR training set, we also prepare a testing set for multi-reference RefSR testing. We remove the images containing target or reference patches that appeared in the training set. From the remaining image pairs, we construct a testing set consisting of 142 groups, each containing a target image and 2$\sim$6 reference images with image side lengths between 800$\sim$1600.

\subsection{Multi-Reference RefSR network}
\label{sec:mrsa}

Armed with the LMR dataset, we propose a multi-reference RefSR network to make good use of multiple reference images, dubbed MRefSR.
Our MRefSR is based on $C^2$-Matching~\cite{jiang2021robust} as it is currently the open source method with best performance and easy to get started.  \zhanglin{Note that other RefSR frameworks such as TTSR~\cite{yang2020learning} are also applicable since we aims to exploit multi-reference features instead of single-reference feature transfer.}
As $C^2$-Matching did, a \emph{Content Extractor} (CE) is used to extract features $F_{LR}$ from $LR$ image. 
Multi-scale ($1\times$, $2\times$ and $4\times$) reference features $F^s_{Ref_{i}}$ are extracted by a \emph{VGG} extractor, where $s = 1, 2, 4$ and $i\in\{1,2,...,N\}$, $N$ is the number of reference images. For the sake of brevity, the $s$ in the following are omitted, and $F_{Ref_{i}}$ is used instead of $F^s_{Ref_{i}}$.
A pretrained \emph{Contrastive Correspondence Network} (CCN) is used to obtain the relative target offsets $O_{i}$ of the LR input and the corresponding multiple reference images. 
Afterwards, as shown in Figure~\ref{fig:Method}, we develop a \textbf{M}ulti-Reference \textbf{A}ttention \textbf{M}odule (MAM) for the multi-reference feature fusion and a \textbf{S}patial \textbf{A}ware \textbf{F}iltering \textbf{M}odule (SAFM) for the fused feature selection.

\emph{Dynamic Aggregation Module} in $C^2$-Matching is used to get the aligned features $F_{a_{i}}$ from the reference features $F_{Ref_{i}}$ by the corresponding pre-offsets $O_{i}$. After that, we introduce MAM to fuse the aligned features from different reference images. 
In detail, at each feature scale, we first generate corresponding $N$ attention maps for the aligned features of $N$ reference images:
\begin{equation}
\label{equa:attention1}
\begin{aligned}
att_{i} (x,y) &= softmax(\left \langle Q(x,y), K_{i}(x,y)\right \rangle) \\
              &= \frac{exp( \left \langle Q(x,y), K_{i}(x,y)\right \rangle )}{\sum_{j=1}^{N} exp( \left \langle Q(x,y), K_{j}(x,y)\right \rangle)}.
\end{aligned}
\end{equation}

We use inner product to measure the similarity between the features $Q(x,y)$ and $K_{i}(x,y)$ at the point $(x, y)$,  where query $Q$ is obtained from the LR input feature $F_{LR}$, key $K_{i}$ and value $V_{i}$ are obtained from the $i$-th reference image aligned feature $F_{a_{i}}$:
\begin{equation}
\label{equa:QKV}
\begin{aligned}
Q &= conv_{q} (F_{LR}), \\
K_{i} &= conv_{k} (F_{a_{i}}), \\
V_{i} &= conv_{v} (F_{a_{i}}), \\
\end{aligned}
\end{equation}
where $conv_{q}$, $conv_{k}$ and $conv_{v}$ are convolutions with kernel size 3$\times$3 and stride 1. Then, we get fused reference feature $F_{fref}$ from all reference images:
\begin{equation}
\label{equa:fref}
\begin{aligned}
F_{fref}(x, y) = \sum_{i=1}^{N} (att_{i}(x,y) \cdot V_{i}(x,y)).
\end{aligned}
\end{equation}

The proposed MAM enables MRefSR to handle an arbitrary number of reference images during training and testing phases, making the MRefSR more flexible for practical applications.

Since not all LR feature pixels can be well matched with reference features, we use the proposed SAFM for the selection of fused reference features $F_{fref}$.
As shown in Figure~\ref{fig:Method}, we get two masks $M_{mul}$ and $M_{add}$ from the concatenated feature of $F_{LR}$ and $F_{fref}$ and a $sigmoid$ function is used to limit the range of the $M_{mul}$.

\begin{equation}
\label{equa:mask}
\begin{aligned}
M_{mul} &= sigmoid(f_{1}(F_{LR}\|F_{fref})) \cdot  2, \\
M_{add} &= f_{2}(F_{LR}\|F_{fref}),
\end{aligned}
\end{equation}
where $f_{1}$ and $f_{2}$ are nonlinear mapping functions consisting of convolution and leaky ReLU layers. At last, the $M_{mul}$ and $M_{add}$ are used for the final selected reference features $F_{sref}$:
\begin{equation}
\label{equa:slt_ref}
\begin{aligned}
F_{sref} = F_{fref} \odot  M_{mul} + M_{add},
\end{aligned}
\end{equation}
where $\odot$ denote element-wise multiplication.

In the end, a restoration module $\mathcal{G}$ takes the LR features $F_{LR}$ and the selected reference features $F_{sref}$ to reconstruct the target image:
\begin{equation}
\label{equa:rec}
X_{SR} = \mathcal{G}(F_{LR}, F_{sref}).
\end{equation}

\subsection{Implementation Details}
We train and evaluate our MRefSR in a scale factor $4\times$. In detail, we train the network for 255K iterations using Adam optimizer~\cite{kingma2015adam} with parameters $\beta_1=0.9$, $\beta_2=0.999$, and constant learning rate of 1e-4. 
Each mini-batch includes 48 groups of image patches, each consisting of an LR input patch with size 40$\times$40 and five reference HR patches with size 160$\times$160. 
We use three commonly used loss functions to train our model, including reconstruction loss $L_{rec}$, perceptual loss $L_{per}$ , and adversarial loss $L_{adv}$, referring to supplementary material for the network training loss details. The weight coefficients for $L_{rec}$, $L_{per}$ and $L_{adv}$ are set to 1, 1e-4 and 1e-6. The network is first trained with $L_{rec}$ only and then finetuned with all losses.
During training, we augment the training data by randomly horizontally flipping and vertically flipping, and random {90\textdegree} rotation. 
Following the standard protocol, we generate all LR images by bicubically downsampling the HR images with a scale factor of $4\times$. All experiments run in parallel on 4 NVIDIA V100 GPUs.
For the quantitative comparison, we train MRefSR without GAN loss and perceptual loss as other methods did. 
Benefiting from the large-resolution training images of LMR, we get an LPF (Large Patch Finetuning) version of the model, which is finetuned using the large-patch training images.
\section{Experiments}
\label{experiments}

\begin{table*}[t]
\centering
\caption{We report PSNR/SSIM on Y channel of YCbCR space to compare among different SR
methods on the testing set of LMR, CUFED5~\cite{zhang2019image,wang2016event}, Sun80~\cite{sun2012super}, and WR-SR~\cite{jiang2021robust}. Methods are grouped by SISR methods (top) and reference-based methods (bottom). The best results are marked \underline{\bf in bold and with underlines}. The second best and the third best results are marked in {\bf bold} and with \underline{underlines}, respectively. $C^2$-Matching-LMR means $C^2$-Matching-$rec$ is trained on the LMR dataset and the Ours-$rec$-LPF indicates that the model was finetuned using large patch size (300$\times$300) training images.}
\label{table:results}  
\resizebox{\textwidth}{!}{%

\begin{tabular}{l|c|c|c|c|c}
\toprule
\multirow{2}{*}{Method} & \multirow{2}{*}{\makecell{Training\\Dataset}} & LMR & CUFED5~\cite{zhang2019image,wang2016event} & Sun80~\cite{sun2012super} & WR-SR~\cite{shim2020robust} \\
\cline{3-6}
& & PSNR$\uparrow$ / SSIM$\uparrow$ & PSNR$\uparrow$ / SSIM$\uparrow$ & PSNR$\uparrow$ / SSIM$\uparrow$ & PSNR$\uparrow$ / SSIM$\uparrow$ \\
\hline
SRCNN~\cite{dong2014learning} & CUFED5 & - & $25.33$ / $0.745$ & $28.26$ / $0.781$ & $27.27$ / $0.767$ \\
EDSR~\cite{lim2017enhanced} & CUFED5 & - & $25.93$ / $0.777$ & $28.52$ / $0.792$ & $28.07$ / $0.793$ \\
RCAN~\cite{zhang2018image} & CUFED5 & - & $26.33$ / $0.781$ & $29.97$ / $0.814$ & $27.91$ / $0.793$ \\
RRDB~\cite{wang2018esrgan} & CUFED5 & - & $26.41$ / $0.783$ & $29.99$ / $0.814$ & $27.96$ / $0.793$ \\
RCAN~\cite{zhang2018image} & LMR & $29.63$ / $0.841$ & $26.58$ / $0.785$ & $\mathbf{30.36}$ / $\mathbf{\underline{0.821}}$ & $28.24$ / $0.798$ \\
RRDB~\cite{wang2018esrgan} & LMR & $29.68$ / $0.842$ & $26.61$ / $0.786$ & $\mathbf{\underline{30.37}}$ / $\mathbf{\underline{0.821}}$ & $28.25$ / $0.798$ \\
\hline
Landmark~\cite{yue2013landmark} & CUFED5 & - & $24.91$ / $0.718$ & $27.68$ / $0.776$ & - \\
CrossNet~\cite{zheng2018crossnet} & CUFED5 & - & $25.48$ / $0.764$ & $28.52$ / $0.793$ & - \\
SRNTT-$rec$~\cite{zhang2019image} & CUFED5 & - & $26.24$ / $0.784$ & $28.54$ / $0.793$ & $27.59$ / $0.780$ \\
TTSR-$rec$~\cite{yang2020learning} & CUFED5 & $29.13$ / $0.832$ & $27.09$ / $0.804$ & $30.02$ / $0.814$ & $27.97$ / $0.792$ \\
MASA-$rec$~\cite{lu2021masa} & CUFED5 & $29.42$ / $0.837$ & $27.54$ / $0.814$ & $30.15$ / $0.815$ & $28.19$ / $0.796$ \\
$C^2$-Matching-$rec$~\cite{jiang2021robust} & CUFED5 & $30.01$ / $0.856$ & $28.40$ / $0.846$ & $30.18$ / $0.817$ & $28.32$ / $0.801$ \\
AMSA-$rec$~\cite{xia2022coarse} & CUFED5 & - & $28.50$ / $0.849$ & $30.29$ / $0.819$ & - \\
TDF-$rec$~\cite{huang2022task} & CUFED5 & - & $28.64$ / $0.850$ & $30.31$ / $\underline{0.820}$ & $\underline{28.52}$ / $\mathbf{\underline{0.807}}$ \\
$C^2$-Matching-LMR & LMR & $\underline{30.64}$ / $\underline{0.869}$ & $\underline{28.65}$ / $\underline{0.853}$ & $30.31$ / $0.819$ & $\mathbf{28.53}$ / $\mathbf{\underline{0.807}}$ \\
Ours-$rec$ & LMR & $\mathbf{31.81}$ / $\mathbf{0.895}$ & $\mathbf{28.94}$ / $\mathbf{0.860}$ & $30.28$ / $0.819$ & $\underline{28.52}$ / {$0.806$} \\
Ours-$rec$-LPF & LMR & $\mathbf{\underline{31.98}}$ / $\mathbf{\underline{0.898}}$ & $\mathbf{\underline{29.05}}$ / $\mathbf{\underline{0.862}}$ & $\underline{30.32}$ / $0.819$ & $\mathbf{\underline{28.59}}$ / $\mathbf{\underline{0.807}}$ \\
\bottomrule
\end{tabular}
}
\end{table*}
\begin{figure*}[!t]
    \centering
	\begin{tabular}{p{0.152\textwidth}<{\centering}p{0.128\textwidth}<{\centering}p{0.145\textwidth}<{\centering}p{0.124\textwidth}<{\centering}p{0.157\textwidth}<{\centering}p{0.135\textwidth}<{\centering}}
		LR input  & Ref-1 & Ref-2 & Ref-3 & Ref-4 & Ref-5 \\
		\hline
		Target HR & ESRGAN & MASA & $C^2$-Matching & $C^2$-Matching-LMR & Ours \\
	\end{tabular}
    \includegraphics[width=0.985\textwidth]{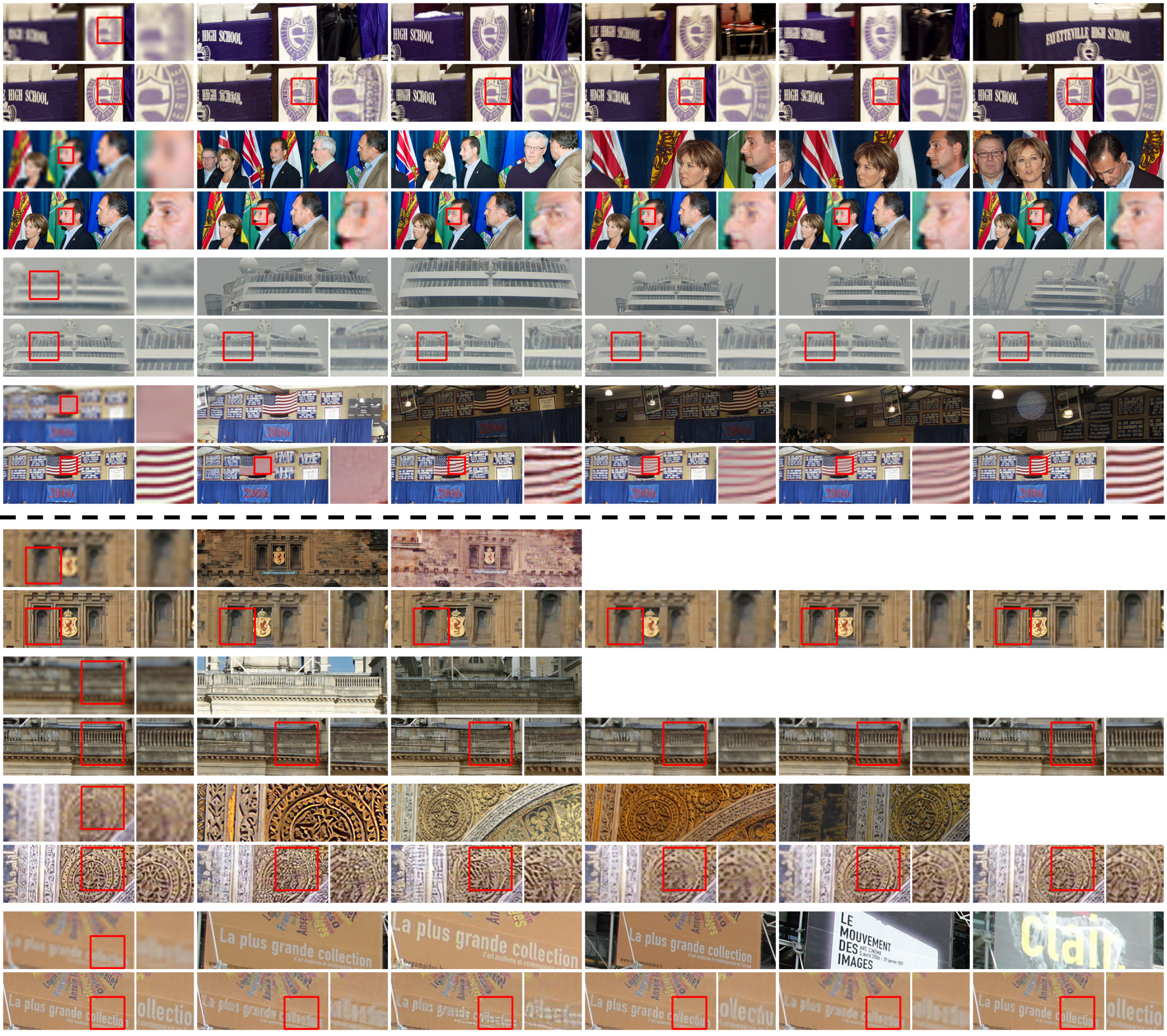}
    \caption{Qualitative comparisons on the testing set of CUFED5 (the top four examples) and LMR (the bottom four examples). We compare our results with ESRGAN, MASA, $C^2$-Matching, $C^2$-Matching-LMR. All these methods are trained with GAN loss. Our method reconstructs sharper details than other methods. }

    \label{fig:results}
\end{figure*}
\begin{figure}[t]
	\centering
	\includegraphics[height=4.95cm]{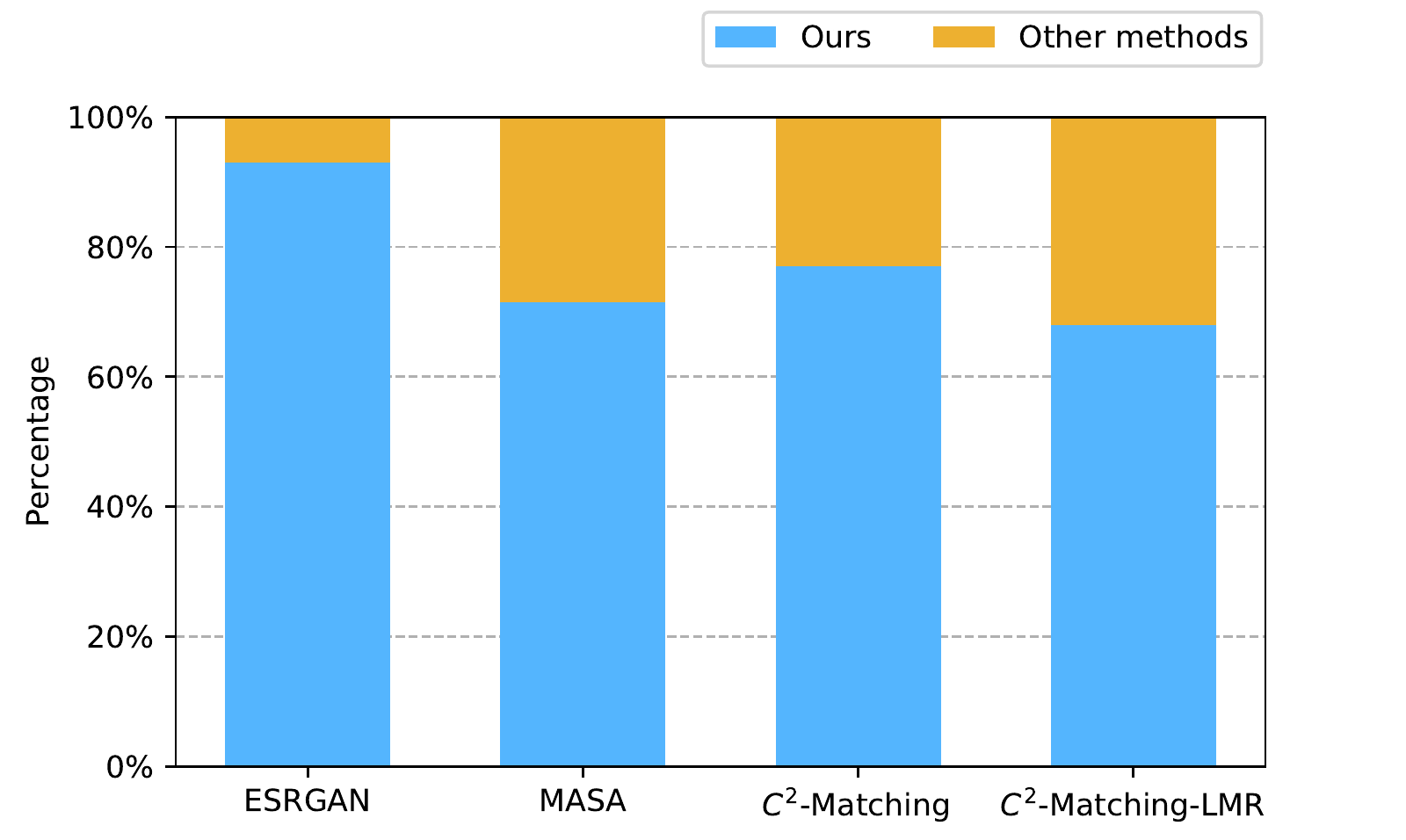}
	\caption{User study results. Values on Y-axis denote the voting percentage of users favoring our method.}
	\label{fig:user_study}
\end{figure}
\begin{figure}[t]
	\centering
	\includegraphics[height=2.49cm]{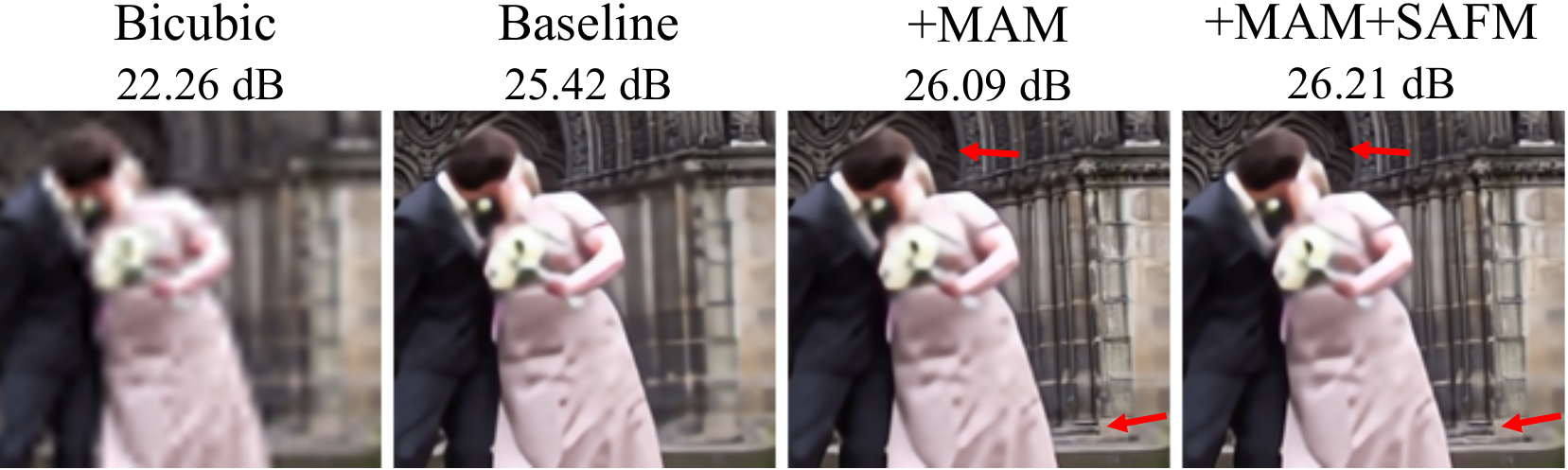}
	\caption{Visual comparisons of ablation study on MAM and SAFM. }
	\label{fig:ablation}
\end{figure}

\subsection{Datasets and Metrics}
We train our network on the proposed LMR training set and evaluate it on the testing set of LMR, CUFED5~\cite{zhang2019image,wang2016event}, Sun80~\cite{sun2012super} and WR-SR~\cite{jiang2021robust}. 
As mentioned earlier, LMR and CUFED5 are two real multi-reference testing sets.
Although Sun80 has multiple reference images, these reference images are not very similar to the corresponding target images.
WR-SR is a single-reference testing set. 
According to previous practice, due to the relatively small images of CUFED5 testing set (300$\times$500), when testing other single-reference RefSR methods on CUFED5, we stitch multiple reference images into one large reference image for testing.
However, on the LMR and Sun80, due to the large image resolution of the testing sets and the limitation of GPU memory, other RefSR methods cannot be tested by the stitching reference image together and can only use a single reference image.
With the multi-reference attention module (MAM), our MRefSR can utilize multiple reference images for prediction on the LMR, CUFED5 and Sun80 testing sets.
we adopt two quantitative metrics, PSNR and SSIM, both calculated on Y channel in the transformed YCbCr color space. 
To evaluate the results qualitatively, we show the visual results of different methods and conduct a user study for subjective visual quality comparison.

\subsection{Comparison with State-of-the-Art Methods}
We compare the proposed MRefSR with previous state-of-the-art SISR methods and single-reference RefSR methods. SISR methods include SRCNN~\cite{dong2014learning}, EDSR~\cite{lim2017enhanced}, RCAN~\cite{zhang2018image}, RRDB~\cite{wang2018esrgan} and ESRGAN~\cite{wang2018esrgan}. As for single-reference RefSR methods, Landmark~\cite{yue2013landmark}, CrossNet~\cite{zheng2018crossnet}, SRNTT~\cite{zhang2019image}, TTSR~\cite{yang2020learning}, MASA~\cite{lu2021masa}, $C^2$-Matching~\cite{jiang2021robust}, 
AMSA~\cite{xia2022coarse} and TDF~\cite{huang2022task} are included. 
For fair comparison, we retrain three high-performance SISR methods RCAN, RRDB and ESRGAN, and one open-sourced top-performing single-reference RefSR method $C^2$-Matching on the training set of LMR.

\textbf{Quantitative evaluation.}
As shown in Table~\ref{table:results}, our MRefSR outperforms other methods by a large margin on two real multiple reference datasets, CUFED5 and LMR. 
On the most commonly used CUFED5 benchmark, MRefSR outperforms the retrained $C^2$-Matching-LMR by 0.29dB. 
Models trained on LMR can achieve better performance on CUFED5, which also demonstrates the generalization ability and effectiveness of LMR.
What's more, MRefSR shows a significant improvement of 1.15 dB over the second best method on the LMR testing set. 
\zhanglin{The above two results demonstrate the superiority of learning the interaction among multiple references, further manifesting the necessity of the LMR dataset that enables multi-reference RefSR training.}
On Sun80, SISR methods RRDB and RCAN get the best two results. The results gap of the top RefSR methods AMSA-rec, TDF-rec, $C^2$-Matching-LMR and MRefSR are less than 0.04 dB, which further proves the reference image and its target image in Sun80 are not very similar.
On the WR-SR benchmark, since there is only one reference image per LR, our results are very close to $C^2$-matching-LMR.

\textbf{Qualitative evaluation.}
As shown in Figure~\ref{fig:results}, we compare the results of ESRGAN, MASA, $C^2$-Matching, $C^2$-Matching-LMR and our MRefSR. 
The top four examples are from CUFED5, and the models trained on LMR generalize well on the CUFED5 testing set, demonstrating the effectiveness of the proposed LMR. 
What's more, the results of our MRefSR trained with multiple references are much better than those trained with a single reference image. 
We also show four examples from the LMR testing set, and MRefSR can recover more texture details than other methods.

Besides, we perform a user study to compare with some typical methods including ESRGAN, MASA and $C^2$-Matching. 
Specifically, in each test, we present paired super-resolution results, one of which is generated by our MRefSR, and ask the users to choose the one with higher visual quality. 
As shown in Figure~\ref{fig:user_study}, the users prefer our results over the others.

\subsection{Ablation Study}
In this section, we verify the effectiveness of Multi-reference Attention Module (MAM) and Spatial Aware Filtering Module (SAFM). Besides, we demonstrate the benefit of large-resolution training images of LMR. At last, we also investigate the impact of number of reference images.

\begin{table}
\caption{Ablation study on the influence of MAM, SAFM and LPF.}
\label{table:ablation_MRefSR}
\centering
\scalebox{0.80}{
\begin{tabular}{l|c|c}
\hline
\multirow{2}{*}{Method} & LMR & CUFED5 \\
\cline{2-3}
& PSNR$\uparrow$ / SSIM$\uparrow$ & PSNR$\uparrow$ / SSIM$\uparrow$ \\
\hline
Baseline($C^2$-Matching-LMR)       & $30.64$ / $0.869$ & $28.65$ / $0.853$ \\
Baseline+MAM                   & $31.70$ / $0.894$ & $28.85$ / $0.859$ \\
Baseline+MAM+SAFM                & $31.81$ / $0.895$ & $28.94$ / $0.860$ \\
\makecell{Baseline+MAM+SAFM+LPF} & $31.98$ / $0.898$ & $29.05$ / $0.862$ \\
\hline
\end{tabular}
}
\end{table}

\textbf{The effectiveness of MAM and SAFM.} As shown in Table~\ref{table:ablation_MRefSR}, with $C^2$-Matching-LMR as the baseline, our MAM achieves a PSNR improvement of 1.06 dB on LMR and 0.20 dB on CUFED5. The reason why the improvement on LMR is larger than that on CUFED5 is that $C^2$-Matching-LMR cannot use multiple reference images on LMR due to the limitation of GPU memory, and MAM greatly solves this problem. 
More importantly, our MAM supports an arbitrary number of reference images, making it more flexible and practical.
On the basis of MAM, SAFM is used to adjust the fused reference features and the PSNR scores on LMR and CUFED5 increase to 31.81 dB and 28.94 dB, respectively.
\zhanglin{Figure~\ref{fig:ablation} shows their influence.}
Furthermore, thanks to the larger image size of the LMR training data, MRefSR with large-patch (300$\times$300) finetuning strategy (LPF) can consistently improve the performance by roughly 0.1 dB on LMR and CUFED5.
This result reflects the advantage of the large training images of the LMR dataset.

\FloatBarrier
\begin{table}[t]
\caption{The effect of different number of reference images on CUFED5.}
\label{table:ablation_nref}
\centering
\resizebox{0.38\textwidth}{!}{
\begin{tabular}{l|c|c}
\hline
$\#Num$ & $C^2$-Matching-LMR & Ours  \\
\hhline{=|=|=}
$n=1$ & $28.474$ & $28.663$ \\
$n=2$ & $28.615$ ($+0.141$) & $28.869$ ($+0.206$) \\
$n=3$ & $28.651$ ($+0.036$) & $28.920$ ($+0.051$) \\
$n=4$ & $28.649$ ($-0.002$) & $28.932$ ($+0.012$) \\
$n=5$ & $28.650$ ($+0.001$) & $28.935$ ($+0.003$) \\
$\;\;\;\Delta$ & $+0.176$ & $+0.272$ \\
\hline
\end{tabular}
}
\end{table}

\textbf{The effect of the number of reference images.} To study the influence of number of reference images, we conduct experiments on the testing set of CUFED5, in which each LR input image has five reference images. 
As shown in Table~\ref{table:ablation_nref}, as the number of reference images increases, although $C^2$-Matching-LMR has a slight improvement with the stitching testing strategy, the gap is still smaller than the improvement of MRefSR. 
What's more, when the number of reference images is greater than 3, the results are worse than the case of 3 reference images, which indicates that 
\zhanglin{the stitching testing strategy neglects the interaction among references, so the information from the fourth reference doesn't explore with that from the first three reference effectively in this case.}
In contrast, it can be seen that with the increase of reference images, MRefSR has a stable positive gain. Last but no least, MRefSR with five references has a PSNR increase of 0.272 dB than that with one reference, whereas $C^2$-Matching-LMR only has a PSNR increase of 0.176 dB, which further demonstrates the superiority of 
\zhanglin{modeling the relationship among multiple references.}

\subsection{Computational Cost}

Here, we present the computational cost comparisons between the proposed MRefSR and previous single-reference RefSR methods, including MASA~\cite{lu2021masa} and $C^2$-Matching~\cite{jiang2021robust}. The computational cost is computed on CUFED5~\cite{zhang2019image,wang2016event} using one NVIDIA V100 GPU. In specific, for the single-reference RefSR methods on CUFED5, we stitch five reference images into a 2500$\times$500 image as the reference image for testing. Certainly, our MRefSR can directly utilize all the reference images for testing. Table~\ref{table:cost} reports the GPU memory, runtime and performance for each method. Our MRefSR consumes the least GPU memory and achieves the best performance with acceptable runtime.

\begin{table}
\caption{Computational cost and performance comparisons on the testing set of CUFED5. $C^2$-Matching-LMR means $C^2$-Matching-$rec$ is trained on our LMR dataset.}
\label{table:cost}
\centering
\resizebox{0.473\textwidth}{!}{%
\begin{tabular}{c|c|c|c|c}
\bottomrule
Model                              & MASA-$rec$ & $C^2$-Matching-$rec$ & $C^2$-Matching-LMR & MRefSR-$rec$ \\
\hhline{=|=|=|=|=}
GPU Memory (GB)     & 21.98    & 8.37    & 8.37    & 3.42 \\
\hline
Runtime (s)        & 0.417    & 2.29    & 2.29    & 0.875 \\
\hline
PSNR$\uparrow$ & $27.54$ & $28.40$ & $28.65$ & $28.94$ \\
SSIM$\uparrow$ & $0.814$ & $0.846$ & $0.853$ & $0.860$ \\
\bottomrule
\end{tabular}
}
\end{table}

\section{Conclusion}
In this paper, we propose a large-scale multi-reference RefSR dataset: LMR. 
Unlike CUFED5, the only training RefSR dataset available before, LMR has 5 reference images for each LR input image.
What's more, LMR contains 112,142 groups of 300$\times$300 training images, 10 times the number of CUFED5, and the image size is also much larger than CUFED5. 
Besides, we propose a new multi-reference baseline RefSR method, named MRefSR. 
We use a multi-reference attention module (MAM) for feature fusion of an arbitrary number of reference images, and a spatial aware filtering module (SAFM) for the fused feature selection. 
\zhanglin{With LMR enabling multi-reference RefSR training, our method effectively models the relationship among multiple references, thus achieving significant improvements over state-of-the-art approaches on both quantitative and qualitative evaluations. And our method solves the mismatch problem of previous methods using a single reference image for training but testing with multiple reference images.}

{\small
\bibliographystyle{ieee_fullname}
\bibliography{iccvbib}
}

\end{document}